%% file: main.tex
\documentclass{article}

\usepackage{arxiv}
\usepackage[numbers]{natbib}

\usepackage[utf8]{inputenc} % allow utf-8 input
\usepackage[T1]{fontenc}    % use 8-bit T1 fonts
\usepackage{hyperref}       % hyperlinks
\usepackage{url}            % simple URL typesetting
\usepackage{booktabs}       % professional-quality tables
\usepackage{amsfonts}       % blackboard math symbols
\usepackage{nicefrac}       % compact symbols for 1/2, etc.
\usepackage{microtype}      % microtypography
\usepackage{lipsum}
\usepackage{graphicx}
\usepackage{caption}
\makeatletter
\def\blfootnote{\gdef\@thefnmark{}\@footnotetext}
\makeatother
\theoremstyle{plain}

\theoremstyle{definition}

\theoremstyle{remark}

\title{Considerations on the Theory of Training Models with Differential Privacy}

\author{Marten van Dijk$^{1,2,3}$ \textit{and} Phuong Ha Nguyen$^{4}$ \\
$^{1}$ CWI Amsterdam, The Netherlands\\
$^{2}$ Department of Computer Science, Vrije Universiteit Amsterdam, The Netherlands \\
$^{3}$ 
Department of Electrical and Computer Engineering, University of Connecticut, CT, USA\\
$^{4}$ eBay, CA, USA\\
\\
\texttt{marten.van.dijk@cwi.nl} and \texttt{phuongha.ntu@gmail.com}}

\begin{document}
\maketitle
\blfootnote{$^{\dagger}$ This is a book chapter}

\begin{abstract}
In federated learning collaborative learning takes place by a set of clients who each want to remain in control of how their local training data is used, in particular, how can each client's local training data remain private? Differential privacy is one method to limit privacy leakage. We provide a general overview of its framework and provable properties, adopt the more recent hypothesis based definition called Gaussian DP or $f$-DP, and discuss Differentially Private Stochastic Gradient Descent (DP-SGD). We stay at a meta level and attempt intuitive explanations and insights \textit{in this book chapter}.
\end{abstract}

% keywords can be removed

\input{Section/BookChapterDP20230702}
\bibliography{references}
\bibliographystyle{plainnat}

\clearpage
\appendix
\onecolumn

\end{document}

%% file: Section/BookChapterDP20230702.tex
\section{Introduction}\label{chapter_XX_intro}

Privacy leakage is a big problem in the big-data era.
Solving a learning task based on big data intrinsically means that only through a collaborative effort sufficient data is available for training a global model with sufficient clean accuracy (utility).
Federated learning is a framework where a learning task is solved by a loose federation of participating devices/clients which are coordinated by a central server \citep{mcmahan,jianmin,bonawitz2019towards,konevcny2016federated,GoogleAIBlog,yang,luping,kevin,cong2019,yang2019,tianli2019hetero,lian2015asynchronous,lian2017asynchronous,ManiaPanPapailiopoulosEtAl2015,DeSaZhangOlukotunEtAl2015,jie2020asynchronous}.
Clients, who use own local data to participate  in a learning task by training a global model, want to have privacy guarantees for their local proprietary data. For this reason DP-SGD \citep{abadi2016deep} was introduced as it adapts distributed Stochastic Gradient Descent (SGD)\citep{van2020hogwild_arxiv} with Differential Privacy (DP)\citep{dwork2006calibrating, dwork2011firm,dwork2014algorithmic,dwork2006our}.

The optimization problem for training many Machine Learning (ML) models using a training set $\{\xi_i\}_{i=1}^m$ of $m$ samples can be formulated as a finite-sum minimization problem as follows
\begin{equation}\label{eq:finite_sum_main}
\min_{w \in \mathbb{R}^d} \left\{ F(w) = \frac{1}{m}
\sum_{i=1}^m f(w; \xi_i) \right\}.
\end{equation}
The objective is to minimize a loss function with respect to model parameters $w$. This problem is known as empirical risk minimization and it covers a wide range of convex and non-convex problems from the ML domain, including, but not limited to, logistic regression, multi-kernel learning, conditional random fields and neural networks.

We want to solve (\ref{eq:finite_sum_main}) in a distributed setting where many clients have their own local data sets and the finite-sum minimization problem is over the collection of all local data sets. A widely accepted approach is to repeatedly  use the SGD \citep{RM1951,nguyen2018sgd,nguyen2018new}
%Stochastic Gradient Descent (SGD)
recursion
\begin{equation}
 w_{t+1} = w_t - \eta_t  \nabla f(w_t;\xi),\label{eqwSGD}
 \end{equation}
 where $w_t$ represents the model after the $t$-th iteration; $w_t$ is used in computing the gradient of $f(w_t;\xi)$, where $\xi$ is a data sample randomly selected from the data set $\{\xi_i\}_{i=1}^m$ which comprises the union of all local data sets.

This approach allows each client to perform local SGD recursions for the data samples $\xi$ that belong to the client's local training data set. The updates as a result of the SGD recursion (\ref{eqwSGD}) are sent to a centralized server who aggregates all received updates and maintains a global model. The server regularly broadcasts its most recent global model so that clients can use it in their local SGD computations. This allows each client to use what has been learned from the local data sets at the other clients. This leads to good accuracy of the final global model.

%==============

%This approach allows each client to perform local SGD computations for the data samples $\xi$ that belong to the client's local training data set $d$. The updates as a result of the SGD computations are sent to a centralized server who aggregates all received updates and maintains a global model. The server regularly broadcasts its most recent global model so that clients can use it in their local SGD computations. This allows each client to use what has been learned from the local data sets at the other clients. This leads to good accuracy of the final global model.

Each client performs SGD recursions for a batch of local data. These recursions together represent a local round and at the end of the local round a local model update (in  the form of an aggregate of computed gradients during the round) is transmitted to the server.
The server in turn adds the received local update to its global model -- and once the server receives new updates from (a significant portion of) all  clients, the global model is broadcast to each of the clients. When considering privacy, we are concerned about how much information these  local updates reveal about the used local data sets. Each client wants to keep its local data set as private as possible with respect to the outside world which observes round communication (the outside world includes all other clients as well).

%==================

%Each client is doing SGD recursions for a batch of local training data. These recursions together represent a local round and at the end of the local round the sum of local model updates, i.e., the addition of computed gradients, is transmitted to the server. The server in turn adds the received sum of local updates to its global model -- and once the server receives new sums from all clients, the global model is broadcast to each of the clients. When considering privacy, we are concerned about how much information these sums of local updates reveal about the used local data sets. Each client wants to keep its local data set as private as possible with respect to the outside world which observes round communication (the outside world includes all other clients as well).

Rather than reducing the amount of round communication such that less sensitive information is leaked, differential privacy \citep{dwork2006calibrating, dwork2011firm,dwork2014algorithmic,dwork2006our} offers a solution in which each client-to-server communication is obfuscated by noise. If the magnitude of the added noise is not too much, then a good accuracy of the global model can still be achieved albeit at the price of more overall SGD iterations needed for convergence.
%achieving good accuracy. 
On the other hand, only if the magnitude of the added noise is large enough, then good differential privacy guarantees can be given. This leads to a friction between desired differential privacy and desired utility/accuracy.

%When using Gaussian based differential privacy, each local round produces a sum of locally computed gradients based on the local data set, which is transmitted to the central server after adding Gaussian noise  \citep{abadi2016deep}. Because of this, each local training round communicates information about the corresponding local data set. Privacy leakage aggregates over multiple training rounds, and the total amount of privacy leakage from the local data set is expected to increase as the total number of local training rounds increases. 

Section \ref{sec:DP-SGD} starts discussing DP-SGD \cite{abadi2016deep}, which implements differentially private mini-batch SGD. Section \ref{sec:DP} explains differential privacy with various (divergence based) measures and properties. Section \ref{sec:GDP} continues detailing the state-of-the-art hypothesis testing based differential privacy, called $f$-DP \cite{dong2021gaussian}, applied to DP-SGD. We conclude with open questions in Section \ref{sec:future}.

%==============

%utility = (test) accuracy, influenced by clipping noise and Gaussian noise -- discuss both, influenced by round frequency

%actual membership attacks?

%==========

%setting: learning problem -- each individual contributes training (and test) data -- the final model is based on the collection of all data

\section{Differential Private SGD (DP-SGD)} \label{sec:DP-SGD}

%\subsection{DP-SGD}
%\label{sec:dpsgd}

We analyse the Gaussian based differential privacy method, called DP-SGD, of \citep{abadi2016deep} 
in a distributed setting with many clients and a central aggregating server. A slightly generalized description of DP-SGD is depicted in Algorithm \ref{alg:DPs}.
The main goal of DP-SGD is to hide whether the collection of transmitted round updates $\bar{U}$ corresponds to data set $d$ versus a neighboring data set $d'$; sets $d$ and $d'$ are called neighbors if they differ in exactly one element. In order to accomplish this, DP-SGD introduces noise, which we will see comes in two flavors {\em clipping noise} and {\em Gaussian noise}.

%between  maximum {\em sensitivity} of $C$, that is, for neighboring data sets $d$ and $d'$ ...
%
%Gaussian DP assumes that all gradients are bounded by some constant $C$ (this is needed in the DP proofs of \citep{abadi2016deep}). However, in general such a bound cannot be assumed (for example, the bounded gradient assumption  is in conflict with strong convexity \citep{nguyen2018sgd}). 
%
%For this reason a constant $C$ is used to clip computed gradients. Once a batch $U$ of gradients is computed, Gaussian noise $n$ is added, after which the result is multiplied by the step size $\bar{\eta}_i$ (and added to the local model $\hat{w}$).
%Experiments in \citep{abadi2016deep} show that such a clipped version of mini-batch SGD  still leads to acceptable convergence to acceptable accuracy. 
%where $E$ is the total number of epochs $E$,

\begin{algorithm}[h]
\caption{Differential Private SGD}
\label{alg:DPs}
\begin{algorithmic}[1]
%\Procedure{LocalSGDwithDP}{$d,C,s,\sigma,T$}
\Procedure{DP-SGD}{}
%    \State \textbf{Input:} dataset (D), batch size (s), number of epochs (K), learning rate ($\eta$), Gradient norm bound (C), noise scale $\sigma$, loss function L, model seed $\hat{w}_0$
    % \State N = len(D) \Comment{Get total number of rows in data set D}
    \State $N=$ size training data set $d=\{\xi_i\}_{i=1}^{N}$
    \State $E=$ total number of epochs
 %   \State $\{C_e\}_{e=1}^{E1}$ sequence of clipping constants (per epoch)
%    \State $T=$ total number of rounds
    \State diminishing step size sequence $\{\eta_i\}$ %_{i=1}^{T}$
    \State 
    \State initialize $w$ as the default initial model
    \State {\bf Interrupt Service Routine (ISR)}: 
  %  \State 
    Whenever a new global model $\hat{w}$ is received, computation is interrupted and an ISR is called that replaces $w\leftarrow \hat{w}$ after which computation is resumed
    \State
    \For{$e\in \{1,\ldots, E\}$}
%    \State re-index data samples: $\{\xi_i\leftarrow \xi_{\pi^e(i)}\}_{i=1}^N$ for a random permutation $\pi^e$
    \State $\{S_{b}\}_{b=1}^{N/m}\leftarrow$ ${\tt Sample}_{m}$ with $S_{b}\subseteq \{1,\ldots, N\}$, $|S_{b}|=m$
    \For{$b\in \{1,\ldots, \frac{N}{m}\}$}
%    \State Set $w$ equal to last received global model
 %   \State $U=0$
 \State Start of round $(e-1)\frac{N}{m}+b$:
    \For{$h\in S_b$}
    \State 
    $a_h= \nabla_w f(w;\xi_h)$
    %$a_h\leftarrow {\cal A}(w,\{\xi_i\}_{i\in S_{b,h}})$
 %   \State $U\leftarrow U+[a]_C$
    \EndFor
    \State $U=\sum_{h=1}^m [a_h]_C$
    \State $\bar{U}\leftarrow U+ {\cal N}(0,(2C\sigma)^2{\bf I})$
    \State Transmit $\bar{U}/m$ to central server
    \State Locally update $w\leftarrow w- \eta_{(e-1)\frac{N}{m}+b} \cdot \bar{U}/m$
    \EndFor
    \EndFor
\EndProcedure
\end{algorithmic}
\end{algorithm}

\subsection{Clipping} \label{sec:clip}

Rather than using the gradient $a_h=\nabla f(w, \xi_h)$ itself, DP-SGD uses its clipped version $[\nabla f(w, \xi_h)]_C$ where 
$$[x]_C= x/\max\{1,\|x\|/C\}.$$ 
%Algorithm \ref{alg:DP} merges the Gaussian Differential Privacy (DP) algorithm of \citep{abadi2016deep} with \Call{LocalSGDwithDP}{}. 
We call this the {\em individual clipping} approach since each computed gradient is individually clipped.
Clipping is needed because 
in general we cannot assume a bound $C$ on the gradients (for example, the bounded gradient assumption  is in conflict with strong convexity \citep{nguyen2018sgd}), yet
the added gradients in update $U$ need to be bounded by some constant $C$ in order for the DP analysis of \citep{abadi2016deep} to go through.
%, see Section \ref{sec:GDP}. 
The reason is that clipping introduces a bound on how much $U=\sum_{h=1}^m [a_h]_C$ gets affected if the differentiating sample between $d$ and $d'$ is used in its computation. Clipping forces a small distance between an update $U$ that does not use the differentiating sample and an update $U'$ that computes the same gradients as $U$ except for one of its gradient computations which
uses the differentiating sample.
%in one of its gradient computations and, otherwise, computes the same gradients as $U$.
%only, by using the differentiating sample, differs in how it is computed. 
This means that if Gaussian noise is added to $U$ and $U'$, respectively, then the smaller the distance between $U$ and $U'$, the harder it is to figure out whether the actually observed noised update originates from $d$ or $d'$. This leads to a differential privacy guarantee.

Suppose that $a_h$ influences another gradient computation, e.g., $a_{h+1}$. Then, if the differentiating sample is used in the computation of $a_h$,  this affects not only $a_h$ but also $a_{h+1}$. Even though both $a_h$ and $a_{h+1}$ will be clipped, this increases the distance between $U$ and $U'$, hence, this weakens the differential privacy. For this reason, the different gradient computations $a_h$ in $U$ should be independent of one another. 
%Computation of a gradient $a_h$ should not involve updating the local model $w$ which is used in a computation of another next clipped output. 
In particular, we do not want to implement classical SGD where the computation of $a_h$ updates the local model $w$ which is used in the next gradient computation $a_{h+1}$. This is the reason for implementing mini-batch SGD where each gradient $a_h$  is computed for the same $w$.

%The DP analysis requires each clipped output to be independent from other clipped outputs. This implies that computation of one clipped output should not involve updating the local model $w$ which is used in a computation of another next clipped output. 

%%%% @@@@@@@@@@@@
Clipping introduces clipping noise  defined as the difference between the clipped gradient $[a_h]_C$ and the original gradient $a_h$. This affects the rate of convergence and leads to clipping bias. If convergence is towards   a (local) minimum $w^*$ of (\ref{eq:finite_sum_main}), then DP-SGD will not get closer and closer to $w^*$, but will converge to some model within a radius around $w^*$, where the radius is 
composed of a clipping bias and a bias as a result of the desired DP guarantee by adding Gaussian noise, and the radius is also at least proportional to the last used step size. 
%\textcolor{red}{(clipping bias showing that convergence is to $w^*$ within a radius composed of a clipping bias and bias as a result of the desired DP guarantee by adding Gaussian noise)} 
%\textcolor{red}{TO DO: 
We notice that clipping noise may not reduce from round to round: Even though $\mathrm{E}_\xi[\nabla_w f(w^*;\xi)]$ tends to zero (since we converge to a local minimum), the expected norm of the gradients $c(w^*)=\mathrm{E}_\xi[\| \nabla_w f(w^*;\xi) \|]$ generally does not converge to zero. For this reason, the clipping constant $C$ must be appropriately set (and cannot be too small). E.g., by using the $C$ of another previous related learning task, or using a public data set for estimating  $c(w^*)$ while executing DP-SGD.
%Notice that in general, once convergence sets in, individual gradients tend to get closer to zero. This means that their norms get smaller $\leq C$. This shows that the clipping noise in updates $U$ becomes very small close to zero. 

%We will see, however, that the clipping constant also plays a role in the Gaussian noise added to updates $U$

\subsection{Mini-Batch SGD}

DP-SGD is constraint to a  mini-batch SGD approach where before the start of the $b$-th local round in epoch $e$ a random min-batch $S_b$ of sample size $|S_b|=m$ is selected out of a local data set $d$ of size $|d|=N$: In the description of Algorithm \ref{alg:DPs} the sampling is done by a sampling procedure ${\tt Sample}_m$ before the start of epoch $e$ for all rounds together. DP-SGD implements {\em subsampling} which chooses a uniformly random subset $S_b\subseteq d$ of size $m$.

The inner loop 
computes $m$ gradients $a_h=\nabla_w f(w;\xi_h)$. Since there are $N/m$ rounds within an epoch, each epoch has (indeed) a total gradient complexity of $N=|d|$. We notice that each gradient is computed based on $w$ which is the last received global model from the server through the interrupt service routine. In the original DP-SGD, a client waits at the start of a round till it receives the global model which includes the aggregated updates of all previous rounds from all clients. The formulation in Algorithm \ref{alg:DPs} allows for asynchronous behavior, including dropped (or reordering of) messages from the server which can lead to a client missing out on receiving global model versions. More importantly, the server may decide to broadcast global models at a lower rate than the rate(s) at which clients compute and communicate their noised round updates. This allows clients with different compute speeds/resources. Also, the rate at which round updates are computed is not restricted by the throughput of broadcast messages from the server to clients (of course, it remains restricted by the network throughput from the clients through aggregation nodes to the server). This implies that  parameter $m$ can potentially be chosen from the whole range $\{1,\ldots, N\}$ including very small $m$ leading to many round updates per epoch or large $m$ leading to only a couple round updates per epoch. We will later discuss the effect of $m$ on convergence and accuracy and DP guarantee.
%At first, this may not seem to be an advantage since the adversary can now observe many more round updates and we may conclude that more privacy is leaked, hence, a smaller number of rounds per epoch is preferred. We will see that this intuition turns out false. 

We notice that too much asynchronous behavior will hurt convergence of the mini-batch SGD approach and may lead to worse accuracy of the final global model. For this reason, before starting a round, a client can check into what extent the recently received global model deviates from the locally kept model. If this gets too far apart or if the last received global model happened too many rounds ago, then the client will want to wait till a new global model is received and the interrupt service routine is triggered. This implements the necessary synchronous behavior with respect to convergence and accuracy.

\subsection{Gaussian Noise}
\label{sec:noise}

The clipped gradients $[a_h]_C$ are summed together in round update $U$.
%maintains the sum $U$ of gradient updates where each of the gradients correspond to the same local model $\hat{w}$ until it is replaced by a newer global model at the start of the outer loop. 
%
At the end of each local round the round  update $U$ is obfuscated by adding Gaussian noise 
$${\cal N}(0,(2C\sigma)^2)$$
to each of $U$'s vector entries. The resulting noised round update $\bar{U}$ divided by the mini-batch size $m$   is transmitted to the server. 
%In this general description $\sigma_i$ is round dependent, but our DP analysis in supplementary material must from some point onward assume a constant $\sigma=\sigma_i$ over all rounds. 
%The noised $U$ times the round step size $\bar{\eta}_i$ is added to the local model after which a new local round starts again.
%
%The noised $U$ is  transmitted to 

For neighboring data sets $d$ and $d'$, we have that the {\em sensitivity} measured as the Euclidean distance between $U$ based on $d$ and $U'$ based on $d'$ (see also Section \ref{sec:clip}) is at most $2C$. An adversary trying to distinguish whether the observed update is from $d$ or $d'$ needs to figure out whether the observation is from 
$$U+{\cal N}(0,(2C\sigma)^2{\bf I}) \ \ \mbox{ or } \ \ U'+{\cal N}(0,(2C\sigma)^2{\bf I}).$$ 
Since $\|U-U'\|\leq 2C$, this is at best (for the adversary) equivalent to hypothesis testing between ${\cal N}(0,(2C\sigma)^2)$ and ${\cal N}(2C,(2C\sigma)^2)$. After dividing by $2C$, this is equivalent to  hypothesis testing 
%between 
%${\cal N}(0,\sigma^2)$ and ${\cal N}(1,\sigma^2)$. 
\begin{equation}
  {\cal N}(0,\sigma^2) \ \ \mbox{ versus } \ \  {\cal N}(1,\sigma^2).
  \label{hyp}
\end{equation}
%\textcolor{red}{formula} 
We see that any differential privacy guarantee for the round update is characterized by $\sigma$. 

%%%%%   @@@@@@@@
The argument above does not depend on the properties of function $f$. In fact, we are free in how we compute the $a_h$ in line 14 of Algorithm \ref{alg:DPs}. These may in themselves depend on more than one sample as long as we start with $w$ for each computation of $a_h$. E.g., the $a_h$ may compute updates coming from a local SGD approach (as used in federated learning), they may be computed according to a mini-batch SGD style approach, or some other momentum based approach.
%\textcolor{red}{and the argument above does not depend on the properties of function $f$}

The attentive reader may notice that the original DP-SGD adds ${\cal N}(0,(C\sigma)^2{\bf I})$, a factor 2 less. This is because its DP analysis and proof assume a slightly different subsampling method. In the original DP-SGD we have that each round selects a random mini-batch of {\em exactly} $m$ samples; this leads to the factor $2$  since $U$ and $U'$ will differ in one gradient, hence, $U-U'$ cancels all gradients except for one in $U$ and one in $U'$, both contributing at most $C$ to the norm $\|U-U'\|$, hence, the factor 2. 

However, the software package Opacus \cite{opacus} implements the sampling of DP-SGD differently: Mini-batches do not have a fixed size, they have a  probabilistic size. For each sample $\xi\in d$, we flip a coin and with probability $m/N$ we add $\xi$ to the mini-batch. This means that the {\em expected} mini-batch size is equal to $m$. As a result, the DP analysis of \cite{abadi2016deep} holds true and the factor $2$ can be eliminated. The reason is that now (in the DP analysis) $U'$ has all the gradients of $U$ together with one extra gradient based on the single differentiating sample between $d$ and $d'$. This implies that all gradients in $U-U'$ cancel except for the one based on the differentiating sample, hence, $\|U-U'\|\leq C$. 

In the above argument, we  {\em assume} that the adversary does not learn the actually used mini-batch size otherwise we will again need the factor $2$ (see also Section \ref{sec:strong}).
The observed scaled noised update $\bar{U}/m$ scales in expectation  with the expected norm of a single computed gradient times the used mini-batch size divided by the expected mini-batch size $m$. This shows how $\bar{U}/m$ depends on the used mini-batch size where,
for large $m$ and $N$, it seems reasonable to assume that the adversary cannot gain significant knowledge about the used mini-batch size from 
$\bar{U}/m$.
%an observed scaled noised update $\bar{U}/m$ (which in expectation scales with the expected norm of a single computed gradient times the used mini-batch size divided by the expected mini-batch size $m$, and  therefore depends on the used mini-batch size).
%(which includes noise ${\cal N}(0,(2C\sigma/m){\bf I})$ as a function of the mini-batch size $m$). 
We conclude that a probabilistic mini-batch size is a DP technique that offers a factor $2$ gain.
This chapter summarizes the $f$-DP framework explained for sampling with fixed mini-batch size leading to the extra 
%includes the 
factor 2 (the probabilistic approach can be added as a complimentary technique).

\subsection{Aggregation at the Server}

The server maintains a global model, which we denote by $\hat{w}$. The server adds to $\hat{w}$ the received scaled noised round update $\bar{U}/m$ after multiplying with the round step size\footnote{The client transmits $(b,e,\bar{U})$ to the server and the server knows an a-priori agreed (with the client) round step size sequence. %$\bar{\eta}_{(e-1)\frac{N}{m}+b}$. 
In practice, the client will only transmit a sparsification or lossy compression of $\bar{U}$ where small entries are discarded.} for round $b$ of epoch $e$, 
$$\eta_{(e-1)\frac{N}{m}+b}$$
(the same as the local model update of $w$ by the client). This allows a diminishing\footnote{
%in practice constant step size since 
%%%%%% @@@@@@@@@@@@@
Due to the added Gaussian noise and clipping bias we can only converge to within some radius around a local minimum $w^*$. Therefore, we may use a diminishing step size that converges to a constant step size equal to the anticipated radius. E.g., as a rule of thumb, after every epoch we evaluate the test accuracy based on a public data set  and if not increasing, then we decrease the step size by 10\%.} step size sequence. 
%\textcolor{red}{footnote: in practice constant step size since due to Gaussian noise and clipping noise/bias we can only converge to within some radius around $w^*$, hence, we may set the step size equal to the anticipated radius}
Notice that dividing by the mini-batch size $m$  corresponds to $U$ representing a mini-batch computation  in mini-batch SGD.
%, where $U$ should be averaged by the mini-batch size $m$. 

Each client will select its own DP posture with own selected parameters $m$, $C$, $\sigma$, and own data set $d$ with its own size $N$. It makes sense for the server to collect the noised round updates from various clients during  consecutive time windows and broadcast updated global models at the end of each window. Rather than adding all the received  $\bar{U}$ within a time window to the global model $\hat{w}$ (after multiplying with the appropriate client-specific step sizes and dividing by the appropriate client-specific mini-batch sizes), the server will add a mix of the various local updates. The mix is according to some weighing vector giving more weight to those clients whom the server judges having `better' training data sets for the learning task
%training problem 
at hand. In federated learning the server will ask for each time window a random subset of clients to participate in the training.
In the above context it makes sense to have the step sizes be diminishing\footnote{Continuing the previous footnote, the central server decides when to reduce the step size based on regularly evaluating the test accuracy, and broadcasts the new step sizes to the clients.} from time window to time window rather than have these be client specific. %\textcolor{red}{even controlled -- if accuracy does not improve over last epoch, then reduce the steps size and broadcast this to the clients}

%waiting time in algorithm

%Likely, the step size is fixed per epoch.
%We implicitly assume that step sizes $\bar{\eta}_{(e-1)\frac{N}{m}+b}$ already discount for averaging by $m$.

%TO DO: probability / mix of updates by various clients

%For round $b$ in epoch $e$, the server receives round updates from many, say $k$, clients. Rather than multiplying each received client noised round update by $\bar{\eta}_{(e-1)\frac{N}{m}+b}$, the server will also divide by the number $k$ of clients from which it received updates before aggregating the result into its global model $\hat{w}$. Due to latency and dropping and reordering of messages in the asynchronous setting, the server may not know whether more updates of previous rounds will be received after broadcasting a new global model. Therefore, the server will use in its averaging an estimate of the number $k$ of clients from whom the server is expected to receive noised round updates for round $b$ in epoch $e$. We notice that distributed SGD is robust against asynchronous behavior to a large extend \cite{us, they}.

\subsection{Interrupt Service Routine}

The interrupt service routine will replace the locally kept model $w$ by a received global model $\hat{w}$. This may happen in the middle of a round. We notice that $\hat{w}$ depends on previously transmitted noised round updates by the client and other clients. We will discuss how each of  these previous noised round updates  have a DP guarantee. By the so-called post-processing lemma, these previously transmitted noised round updates can participate in the current computation of a round update $U$ through its dependency  on the global model $\hat{w}$ (through the gradients in $U$) without weakening the DP guarantee for $\bar{U}$ (which includes Gaussian noise on top of $U$). 

Similarly, the client locally updates model $w$ with $\bar{U}$ at the end of a round. In  next rounds this implies that $w$ still only depends on previously transmitted noised round updates by the client and other clients, and again by the post-processing lemma the DP guarantees of future noised round updates do not degrade. As soon as a new global model $\hat{w}$ is received by the interrupt service routine it will overwrite $w$, that is, the current local model is discarded. This is justified because the newly received global model includes the client's own previously communicated noised updates $\bar{U}$ (if the corresponding messages were not dropped and did not suffer too much latency), hence, the information of its own local updates is 
%included 
incorporated in the newly received $\hat{w}$. 
%The difference is that $\hat{w}$ includes averaging by $k$, the number of clients that contributed a noised update to the same round in the same epoch ....

%each client has its own $d$, own $N$, we have different epoch sizes per client ... concept of $k$ is vaque .. do not reveal N to adversary ... likely stepsize fixed for an average epoch size

%by introducing this IRS we make DP-SGD adaptable to the FL setting -- in general, only a subset of clients will contribute each time to a global round update (randomly selected -- discounted in the asynchronous setting, clients choose themselves when and at which rate to contribute)

%may be client step size mu in algo. server decides own step size and averaging etc.

%explain why mini-batch is inevitible -- cannot use classical SGD.

%We call (\ref{indclip}) the {\em individual clipping} approach and  notice that it implements mini-batch SGD. This is necessary because 

%the DP analysis requires each clipped output to be independent from other clipped outputs. This implies that computation of one clipped output should not involve updating the local model $w$ which is used in a computation of another next clipped output. 

%This means that each of the clipped gradients in the sum are evaluated in the same $w$ (the most recently received global model from the server). The main disadvantage of mini-batch SGD is that it does not converge as fast as other SGD-based algorithms such as classical SGD. It is not the individual clipping (with a fine-tuned clipping constant $C$) but mainly the mini-batch SGD approach that makes convergence slow. 

\subsection{DP Principles and Utility}
\label{sec:princ}

The strength of the resulting DP guarantee depends on how much utility we are okay with sacrificing. 
The differential privacy guarantee is discussed in Section \ref{sec:GDP}.
%%% REMOVED %%%
%and turns out to be approximately equivalent to differentiating between samples from ${\cal N}(0,\sigma^2)$ and ${\cal N}(\sqrt{E},\sigma^2)$ (we will also discuss group privacy where $d$ and $d'$ differ in a group of $g$ samples, which will have another $\sqrt{g}$ dependency). 
%This shows that $\sigma$ should be large enough in order to make hypothesis testing between the two normal distributions unreliable. The reason why we have this DP guarantee is because 
The principle of using Gaussian noise {\em bootstraps} DP for each round, see (\ref{hyp}); 
%\textcolor{red}{see formula}; 
the principle of subsampling in the form of random mini-batches of size $m$ {\em amplifies} DP (because only with probability $m/N$ a round uses the differentiating sample and can leak privacy in the first place); and the principle of {\em composition} of DP guarantees for each round over multiple epochs yields the overall DP guarantee.

%In Section \ref{sec:GDP} we discuss DP-SGD's differential privacy posture in $f$-DP language. It turn out that  DP-SGD is $\approx G_{\sqrt{gE}/\sigma}$-DP for group privacy of $g$ samples (i.e., data sets $d$ and $d'$ have $g$ differentiating samples, $d$ and $d'$ are neighboring if $g=1$) and a total of $E$ epochs total gradient computation.

Utility is measured in terms of the (test) accuracy of the final global model and secondary metrics are convergence rate, round complexity $(N/m)\cdot E$ calculated as the total number of rounds per client (communication is costly), total gradient complexity $E\cdot N$ calculated as the total number of computed gradients per client, information dispersal characterized by the delay or latency of what is learned from local data sets which is calculated as the number $m$ of gradient computations between consecutive round communications to the server, and client's memory usage. 
%\footnote{The mini-batch computation $U=\sum_{h=1}^m [a_h]_C$ needs to keep track of all $m$ gradients $a_h=\nabla_w f(w;\xi_h)$, while the unclipped original mini-batch SGD can keep track of $\sum_{h=1}^m a_h = \nabla_w \sum_{h=1}^m f(w;\xi_h)$, a single gradient computing thread.\textcolor{red}{discussion from rebuttal}}. 

The final accuracy depends on the amount of clipping noise and Gaussian noise:
%and also depends on the amount of delay (information dispersal) introduced by $m$:
Once convergence sets in, the clipping noise will be small and close to zero if the clipping constant is appropriately chosen (at least a factor larger than $c(w^*)$, see Section \ref{sec:clip}). However, each round update $U$ has noise sampled from ${\cal N}(0,(2C\sigma)^2{\bf I})$ added to itself. If this noise is small relative to the norm of $U$, then we expect accuracy not to suffer too much if the neural network model is sufficiently robust against noise (it turns out that deeper neural networks are quite sensitive).
%are less robust against (small) noise -- real friction} 
When convergence progresses and the clipping constant is large enough, then $U/m$ behaves like an average of unclipped gradients which is an estimate of 
$\mathrm{E}_\xi[\nabla_w f(w^*;\xi)]$ which tends to zero.
%the gradients in $U$ get closer to zero \textcolor{red}{norms closer to $c(w^*)$} and therefore the norm of $U$ gets smaller, which 
This means that the Gaussian noise relative to the norm of $U$ becomes larger, which puts a limit on how much accuracy/utility can be achieved. 

Since the DP guarantee depends on $\sigma$ but not on $C$ while the added Gaussian noise scales with $C\cdot \sigma$, we will want to implement a form of (differential private) adaptive clipping (we notice that the DP analysis of DP-SGD holds for clipping constants $C$ that vary from round to round).
%
%where $C$ is reduced \textcolor{red}{to $c(w^*)$} when convergence progresses (we notice that the DP analysis of DP-SGD holds for clipping constants $C$ that vary from round to round). 
%This will allow us to contain the Gaussian noise relative to the norm of $U$ when convergence sets in.
Experimentation is needed to fine-tune the parameters $m$, (adaptive) $C$, and $\sigma$.
Despite fine tuning, we remark that  the added clipping and Gaussian noise for differential privacy results in convergence to a final global model with smaller (test) accuracy (than what otherwise, without DP, can be achieved).

%\textcolor{red}{rewrite -- start with thought experiment which is about robustness wrt noise; DP does depend on $m$, we want it small. Also small $m$ gives better accuracy. (say typically $m=64$) The thought experiment shows the complexity of choosing the right $(m,C,\sigma)$.}
%The $\approx G_{\sqrt{gE}/\sigma}$-DP  guarantee for group privacy of Section \ref{sec:GDP} does not reflect the role of the batch size $|S_b|=m$. This is implicitly captured in $\sigma$. 

%%%%% @@@@@@@@@@
The following thought experiment shows how the batch size $m$ influences utility and differential privacy: 
Suppose we increase $m$ to $am$, a factor $a$ larger. Then the norm of updates $U$ will become a factor $a$ larger. As a result, with respect to convergence to the final global model, we should be able to cope with a factor $a$ larger Gaussian noise.
%-- let's make this assumption for now.
That is, by keeping the relative amount of noise with respect to the norm of $U$ constant, the new updates corresponding to batch size $am$ can be noised with 
%$$a\cdot {\cal N}(0,(2C\sigma)^2{\bf I})={\cal N}(0, (2C  \sigma\cdot \sqrt{a})^2{\bf I}).$$ 
$$a\cdot {\cal N}(0,(2C\sigma)^2{\bf I})={\cal N}(0, (2C  \sigma\cdot a)^2{\bf I}).$$ 
In fact the communicated averaged noised round update $\bar{U}/(am)$ has noise 
%$$a\cdot {\cal N}(0,(2C\sigma)^2{\bf I})/(am)={\cal N}(0, (2C \sigma/\sqrt{m})^2{\bf I}),$$
$$a\cdot {\cal N}(0,(2C\sigma)^2{\bf I})/(am)={\cal N}(0, (2C \sigma/m)^2{\bf I}),$$
the same as the original communicated averaged noised round update (before the thought experiment).
This shows that we can use the factor 
%$\sqrt{a}$
$a$ for increasing (1) the clipping constant $C$ (which reduces the clipping noise, which is most prevalent at the start of DP-SGD, so that convergence can more easily bootstrap) and/or increasing (2) the standard deviation $\sigma$ (which improves the DP guarantee);
%$G_{\sqrt{gE}/\sigma}$-DP guarantee as it gets closer to $G_0$ for larger $\sigma$); 
the resulting new clipping constant $C'$ and standard deviation $\sigma'$ satisfy %$2C'\sigma'=2C\sigma\cdot \sqrt{a}$.
$2C'\sigma'=2C\sigma\cdot a$.
%
%A main disadvantage of increasing the batch size with a factor $a$ in combination with increasing the standard deviation of the Gaussian noise with $\sqrt{a}$ is that each 
%
%[TO DO: FINAL ROUND HAS TOO MUCH NOISE -- MUST BE TAPERED DOWN by making $a$ smaller again near the end -- AS SUCH THOUHT EXPERIMENT DOES NOT HOLD, DP SGD also uses limited sigma; need to adjust $a$ to smaller in the final stage]
%

However, the disadvantage of increasing the batch size with a factor $a$ is a multiplicative factor $a$ increased amount of gradient computations since overall we will still need the same number of rounds for convergence, or equivalently, the same number of SGD update steps toward a local minimum $w^*$. This means a factor $a$ larger number of epochs (one epoch measures $N$ gradient computations, hence, if $m$ is increased to $ma$, we have a factor $a$ smaller number of rounds per epoch). But this has a direct impact on the DP guarantee. As we will see in Section \ref{sec:strong} (when discussing Gaussian DP), before the thought experiment we have a $G_\mu$-DP guarantee, where $\mu$ is proportional to $\sqrt{Em/(N\sigma^2)}$ for `large $N$ and $E$.' The thought experiment increases $m$ by $a$ and increases $E$ by $a$. Hence, for the same $\sigma$ we will now have the significantly worse $\approx G_{\mu\cdot a}$-DP guarantee. If the factor $a$ is fully used for increasing $\sigma$ by a factor $a$, then for  `large $N$ and $E$'
%for a 'large number of rounds per epoch'  
the Gaussian DP parameter $\mu\cdot a$ decreases back to $\mu$ and
%(as we will see when discussing Gaussian DP) and 
the overall  DP guarantee remains the same. 
%However, a `large number of rounds' forces a limit to $a$. 
Notice that the total number of gradient computations increases from $E$ to $aE$, while the balance between utility and differential privacy seems to remain\footnote{Here, we notice that a larger $a$ does have the advantage that $h(a\sigma)$ in Section \ref{sec:strong} tends to $1/(a\sigma)$ leading to $\mu$ being proportional to $1/(a\sigma)$ as stated. If $a$ and as a result $a\sigma$ remains relatively small, then $h(\sigma)$ behaves exponentially small in $a\sigma$ and $\mu$ is proportional to $e^{1/(2(a\sigma)^2)}$ which is a much worse dependency on $a\sigma$ leading to an unacceptable DP guarantee.} the same.
The above thought experiment shows that finding the right hyper parameter setting is not straight forward. 

%=====

%The disadvantage of increasing the batch size with a factor $a$ is a multiplicative factor $a$ increased delay, i.e., the number of gradient computations between successive round communications to the server is multiplied by a factor $a$, and this reduces information dispersal and may hurt convergence of the global model. Here, we note that mini-batch SGD is rather robust with respect to large delays, but experiments need to show into what extent $m$ can be increased without affecting the accuracy of the final global model too much. 
%of the Gaussian noise per round update (which directly influences the accuracy of the final global model).

Hyperparameter search depends on the used data set. Either we adopt a hyperparameter setting from another similar learning task, or we search for hyperparameters based on the client data sets. 
In practice, in order to find good parameters $m$, $C$, and $\sigma$, we basically do a grid search by (1) fixing some standard settings (from similar learning tasks) for sample size $m$, e.g., 16, 32, 64, 128
and 256 etc., (2) fixing some standard settings (from similar learning tasks) for clipping constant $C$, e.g., 0.001, 0.01, 0.1, etc., and then (3)
trying some reasonable settings for $\sigma$ (based on the client data sets).
If the grid search indeed uses client data sets, then
%In the latter case, 
we need to make sure that the additional privacy leakage due to the search is small. This is discussed in Appendix D of \cite{abadi2016deep}, see also \cite{Gupta}.

%\subsection{Local SGD}

%\textcolor{red}{ The argument leading to the hypothesis problem of a single round does not use what values are clipped, as long as they are clipped. write about "mini-batch SGD" replacing $\nabla f$ by some other function, like $\xi$ represents a mini-set of samples used to compute an average or classical SGD -- leads to the Local SGD approach in FL, where the whole mini-batch represents one mini-set. }

%We notice that 

%For starting convergence: Since $U$ is the sum of $m$ clipped gradients, $\|U\|\leq mC$ and the noise standard deviation  to norm ratio is at best as small as $2\sigma/m$. The larger $m$, the better. 

%===========

%The effect of parameter $m$, what happens if factor $a$ larger ... eprint discussion.

%thought experiment: a mini-batches together allows a larger noise, hence, sqrt a larger sigma. Better DP guarantee -- or exchange !

%The resulting DP guarantee and its analysis 

%learning task

\subsection{Normalization}

In practice we will also want to use data normalization~\citep{Starovoitov2021DataNI} as a pre-processing step. This requires computing the mean and variance over all data samples from $d$. This makes normalized data samples depend on all samples in $d$. For this reason we need differential private data normalization. That is, a differential private noisy mean and noisy variance is revealed a-priori. This leads to some privacy leakage. The advantage is that we can now rewrite ${\cal A}$ as an algorithm that takes as input $w$, the original data samples $\{\xi_h\}_{h\in S_{b}}$ together with the revealed noisy mean and noisy variance. ${\cal A}$ first normalizes each data sample after which it starts to compute gradients etc. In the $f$-DP framework, privacy leakage is now characterized as a trade-off function of the differential private data normalization pre-processing  composed with the trade-off function corresponding to the DP analysis of DP-SGD (which does not consider data normalization).

We notice that batch normalization is not compatible with the DP analysis of DP-SGD with its individual clipping of each gradient (since this introduces dependencies among the clipped gradients in $U$ and the upper bound of $2C$ on the sensitivity does not hold).  On the other hand layer normalization as well as group and instance normalization are compatible (because these only concern single gradient computations). We notice that if $a_h$ is computed itself by using a local mini-batch SGD approach, then batch normalization  of the used mini-batch can be integrated within its computation. 
%\textcolor{red}{it is compatible with the local mini-batch SGD approach}

As a final remark, our discussion assumes that we already know how to represent data samples by extracting features. We can use Principal Component Analysis (PCA) for dimensionality reduction, that is, learning a set of features which we want to use to represent data samples. PCA can be made differentially private \cite{chaudhuri2013near} in that the resulting feature extraction method (feature transform) has a DP guarantee with respect to the data samples that were used for computing the transform. DP-SGD can be seen as a post-processing after PCA, which is used to represent the local training data samples for which DP-SGD achieves a DP guarantee. In practice, we often already know how to represent the data for our learning task and we already know which function $f(w;\xi)$ to use, i.e., which neural network topology and loss function to use (due to the success of transfer learning we can adopt data representations and $f$ from other learning tasks).

\section{Differential Privacy} \label{sec:DP}

In order to prevent data leakage from  inference attacks in machine learning \citep{lyu2020threats} such as the deep leakage from gradients attack
\citep{ligengzhu201deepleakage,zhao2020idlg,geiping2020inverting} or the membership inference attack
\citep{shokri2017membership,nasr2019MIA,song2019MIA} a range of privacy-preserving methods have been proposed. 
Privacy-preserving solutions 
for federated learning are Local Differential Privacy (LDP) solutions \citep{abadi2016deep,abhishek2018privateFL,mohammad2020privateFL,stacey2018privateFL,meng2020privateFL,duchi2014local} and Central Differential Privacy (CDP) solutions \citep{mohammad2020privateFL,robin2017privateFL,mcmahan2017learning,nicolas2018privateFL,Yu2019CDP}. 
In LDP, the noise for achieving differential privacy is computed locally at each client and is added to the updates before sending to the server -- in this chapter we only consider LDP. In CDP, a {\em trusted server} (aka trusted third party) aggregates received client updates into a global model; in order to achieve differential privacy the server adds noise to the global model before communicating it to the clients.

Differential privacy \citep{dwork2006calibrating, dwork2011firm,dwork2014algorithmic,dwork2006our}, see \cite{BookDP} for an excellent textbook, defines privacy guarantees for algorithms on databases, in our case a client's sequence of mini-batch gradient computations on his/her training data set. The guarantee quantifies into what extent the output  of a client (the collection of updates communicated to the server) can be used to differentiate among two adjacent training data sets $d$ and $d'$ (i.e., where one set has one extra element compared to the other set). 

\subsection{Characteristics of a Differential Privacy Measure}

%\vspace{2mm}
%
%\noindent 
%{\bf Characteristics of a Well Defined Differential Privacy Metric:} 
In DP-SGD, the client wants to keep its local training data set as private as possible. Each noised round update $\bar{U}$ leaks privacy. Let us define round mechanism ${\cal M}_b$ as the round computation that outputs $\bar{U}$ for round $b$. The input of ${\cal M}_b$ is data set $d$ together with an updated local model $w$. We have the following recursion
$$ \bar{U}_b \leftarrow {\cal M}_{b}(w_b;d),
$$
where
$w_b$ is a function of received global model updates which themselves depend on other client's round updates in combination with own previously transmitted round updates $\bar{U}_1,\ldots, \bar{U}_{b-1}$. To express this dependency, we use the notation
$$ w_b \leftarrow {\sf W}(\bar{U}_1,\ldots, \bar{U}_{b-1}),
$$
where ${\tt W}$ receives the global models of the server (and in essence reflects the interrupt service routine).
We define the overall mechanism ${\cal M}$ as the (adaptive) composition of all round mechanisms ${\cal M}_b$, i.e.,
$$
\{\bar{U}_b\} \leftarrow {\cal M}(d) 
\mbox{ with }
 \bar{U}_{b} \leftarrow {\cal M}_{b}({\tt W}(\bar{U}_1,\ldots, \bar{U}_{b-1}) ;d).
$$

When defining a DP measure, we will want to be able to {\em compose} the DP guarantees for the different round mechanisms ${\cal M}_b$: If we can prove that ${\cal M}_b({\tt aux};\cdot)$ has a certain DP guarantee, denoted by ${\tt DP}_b$, for {\em all} ${\tt aux}$ (that can be output by ${\tt W}(\ldots)$), then 
the composition ${\cal M}$ of all round mechanisms ${\cal M}_b$ should have a composed DP guarantee 
$${\tt DP}_1\otimes {\tt DP}_2 \otimes \ldots \otimes {\tt DP}_{(N/m)\cdot E}$$
for some composition tensor $\otimes$ over DP measures.

Once a DP guarantee for mechanism ${\cal M}$ is proven, we do not want it to weaken due to {\em post-processing} of the output of ${\cal M}$. In particular, the central server uses the output of ${\cal M}$ for keeping track of and computing a final global model for the learning task at hand. This final model should still have the same (or stronger) differential privacy posture. Let us denote the post-processing by a procedure ${\tt P}$. If ${\cal M}$ has DP guarantee ${\tt DP}$, then we want ${\tt P}\circ {\cal M}$ to also have DP guarantee ${\tt DP}$ (this is called the post-processing lemma),
$$ [{\tt DP} \mbox{ for } {\cal M}] \ \Rightarrow \ [{\tt DP} \mbox{ for } {\tt P}\circ {\cal M}].
$$

%In order to get a good DP guarantee we want to use subsampling. 
We want our DP measure to be compatible with {\em subsampling}: We want to be able to show that if a round mechanism ${\cal M}_b$ has guarantee ${\tt DP}$ without subsampling, then ${\cal M}_b\circ {\tt Sample}_m$ has an `easy' to characterize amplified guarantee ${\tt DP}'$, `${\tt DP}'\geq {\tt DP}$.'

Finally, we want a differential privacy measure which fits our intuition, in particular, how privacy should be characterized and in what circumstances  an attacker can learn private information from observed mechanism outputs. Differential privacy measures are about the difficulty of distinguishing whether the observed output $o$ is from the distribution  ${\cal M}(d)$ or from the distribution ${\cal M}(d')$, where $d$ and $d'$ are neighboring data sets in that they have all but one differentiating sample in common. The DP guarantee measures in to what extent
$$ {\tt Pr}[o\sim {\cal M}(d)] \ \  \mbox{ and } \ \  {\tt Pr}[o\sim {\cal M}(d')]
$$
are alike for {\em all} neighboring $d$ and $d'$. Here, we want to reflect the intuition that for more likely observations $o$ the two probabilities should be close together while for unlikely observations $o$ we care less whether the two probabilities are close. 
This reflects how we think about the adversary: Only in rare unlikely cases, a lot or all privacy may leak, while in the common case there is very little privacy leakage. 
In cryptology we would want to interpret `rare' as a negligible probability in some security parameter and in the common case we want the two probabilities/distributions to be `statistically close' with their distance negligible in some security parameter. Such strong guarantees cannot be extracted from DP analysis where we control privacy leakage in exchange  for utility/accuracy; we cannot make privacy leakage negligible.

%group privacy

The DP measure is characterized in terms of probabilities and statistics. This is referred to as static security or information theoretical security and allows an adversary with unbounded computational resources in order to differentiate between the hypothesis $o\sim {\cal M}(d)$  and hypothesis $o\sim {\cal M}(d')$. For completeness, in cryptology we also have the notion of computational security meaning that the difficulty of differentiating the two hypotheses can be reduced to solving a computational hard problem (and, since the brightest mathematicians and computer scientists have not been able to find an algorithm which solves this problem efficiently with practical computational resources, we believe that the attacker cannot solve this problem in feasible time). Computational security allows one to obtain security guarantees where the attackers advantage or success is negligible in some security parameter. 
%It is not known how to design computational secure DP with cryptography guarantees.

The above expresses individual privacy. We can generalize towards group privacy by considering data sets $d$ and $d'$ that differ in at most $g$ samples. In this case we say that a mechanism has a DP guarantee with respect to a group of $g$ samples.

%static security (unbounded adversary) in terms of statistics/probabilities

%we cannot guarantee negligible leakage ... we can only control this by setting parameters achieving a strong enough DP for our purposes/requirements, client is in control of its own posture

%inform differential privacy 

%What makes a good DP theory? A measure, composition, postprocessing. Measure has disadvantage -- just neighboring data sets (what about group privacy --- linear scaling in a coarse derivation) and what about the eps and delta, these are large for a cryptographer's taste ..

%==============

%only neighboring sets -- group privacy? / metric has no cryptographical strength

\subsection{$(\epsilon,\delta)$-Differential Privacy} \label{section:epsdelta}

%\vspace{2mm}

%\noindent 
%{\bf $(\epsilon,\delta)$-Differential Privacy:}
%\begin{defn} \label{defDP} 
A randomized mechanism ${\cal M}: D \rightarrow R$ is $(\epsilon, \delta)$-DP (Differentially Private) \cite{dwork2006our} if for any adjacent $d$ and $d'$ in $D$ and for any subset $S\subseteq R$ of outputs,
\begin{equation} {\tt Pr}[{\cal M}(d)\in S]\leq e^{\epsilon}\cdot {\tt Pr}[{\cal M}(d')\in S] + \delta, \label{eq:edDP}
\end{equation}
where the probabilities are taken over the coin flips of mechanism ${\cal M}$.
%\end{defn}

Historically, differential privacy was introduced \cite{dwork2006our} and first defined as $\epsilon$-DP \cite{dwork2006calibrating} which is $(\epsilon,\delta)$-DP with $\delta=0$. In order to achieve $\epsilon$-DP even an unlikely set $S$ of outputs needs to satisfy (\ref{eq:edDP}) for $\delta=0$. This means that the tail distributions of   ${\tt Pr}[{\cal M}(d)\in S]$ and ${\tt Pr}[{\cal M}(d')\in S]$ cannot differ more than a factor $e^\epsilon$. This is a much too strong DP requirement, since the probability to observe an output that corresponds to unlikely tail events is already very small to begin with. Therefore, $\delta$ was introduced so that tail distributions with probability $\leq \delta$ do not need to be close together within a factor $e^\epsilon$. This allows one to achieve the more relaxed $(\epsilon,\delta)$-DP guarantee where an $\epsilon$-DP guarantee cannot be proven.

%====================
%
%A popular relaxation, “approximate” or (ε,δ)-differential privacy [DKM+06], roughly guarantees that with probability at least 1 −δ the privacy loss does not exceed ε. This relaxation allows a δ probability of catastrophic privacy failure, and thus δ is typically taken to be “cryptographically” small. Although small values of δ come at a price in privacy, the relaxation nevertheless frequently permits asymptotically better accuracy than pure differential privacy (for the same value of ε). 
%
%========================

The privacy loss incurred by observing an output $o$ is given by
\begin{equation} L^{o}_{{\cal M}(d) \| {\cal M}(d')} = \ln \left( \frac{{\tt Pr}[{\cal M}(d)=o]}{{\tt Pr}[{\cal M}(d')=o]} \right).
\label{eqloss}
\end{equation}
As explained in \citep{dwork2014algorithmic} $(\epsilon, \delta)$-DP ensures that
for all adjacent $d$ and $d'$ the absolute value of privacy loss will be bounded by $\epsilon$ with probability at least $1-\delta$ (with probability at most $\delta$, observation $o$ is part of the tail); $(\epsilon,\delta)$-DP allows a $\delta$ probability of `catastrophic privacy failure' and from a cryptographic perspective we want this negligibly small. However, when using differential privacy in machine learning we typically use $\delta=1/N$ (or $1/(10N)$) inversely proportional with the data set size $N$ (this seems to correspond well with the intuition when a local update should cause an unlikely/tail observation due to the nature of the specific batch of local data samples that was used in the computation of the local update).
Concerning parameter $\epsilon$,
the larger $\epsilon$ the more certain the adversary is about which of $d$ or $d'$ caused observation $o$.

Compared to $(\epsilon,0)$-DP, the relaxation by $\delta$ allows
%he central advantage of relaxing the guarantee is that it permits 
an improved and asymptotically tight analysis of the cumulative privacy loss incurred by composition of multiple differentially private mechanisms;  \cite {CDP} states an advanced composition theorem (a factor half improvement over \cite{DRV10}):
For all $\epsilon, \delta, \delta'\geq 0$, the class of $(\epsilon,\delta')$-DP mechanisms satisfies $$(\sqrt{2k\ln(1/\delta)}\cdot \epsilon + k\epsilon(e^\epsilon -1)/2, k\delta' +\delta)\mbox{-DP} $$
under $k$-fold adaptive composition. This means that only for $k\leq (1-\delta)/\delta'$ the privacy failure probability remains bounded to something smaller than 1. 

%For example, composition of the $k=(N/m)\cdot E$ round mechanisms ${\cal M}_b$ leads to a $(N/m)\cdot E\cdot \delta'$ term which is both {\em linear} in the number of epochs $E$ and the number of rounds $N/m$ per epoch. As we will see, the advanced composition theorem is not tight for DP-SGD. In fact the $N/m$ dependency can be removed all together. 

%CHECK: plug in parameters, what is the dependency in delta??

%The situation for $(\epsilon,\delta)$-differential privacy is not quite so elegant: 
For group privacy, the literature shows $(g\epsilon,g e^{g-1}\delta)$-DP for groups of size $g$. Here, we see an exponential dependency in $g$ due to the $g e^{g-1}$ term in the privacy failure probability.  This means that only for very small $\delta$, the failure probability remains bounded to something smaller than $1$.
%a troubling exponential increase in the failure probability (the last term).

We conclude that $k$-fold composition and group privacy for group size $g$ only lead to useful bounds for relatively small $k$ and $g$. If we restrict ourselves to a subclass of mechanisms, then we may be able to prove practical DP bounds for composition and group privacy for much larger and practical $k$ and $g$. We will define such subclasses by 
%constraining the privacy loss, that is,
imposing properties on the privacy loss.

%Again, for DP-SGD we will see that we can achieve much tighter bounds.

\subsection{Divergence Based DP Measures}

In order to get better trade-offs for composition and group privacy we want to weigh the tail distribution of unlikely observations in such a way that more unlikely observations are allowed to leak even more privacy. So, rather than weighing all unlikely observations equally likely, which results in the privacy failure probability $\delta$, we want to be more careful. This will allow  improved DP bounds for composition and group privacy.

The first idea is to treat the loss function (\ref{eqloss}) as a random variable $Z$ and note that in a $k$-fold composition we observe $k$ drawings of random variable $Z$. Due to the law of large numbers, the average of these drawings will be concentrated around the mean of the loss function. This leads to the notion of Concentrated Differential Privacy (CDP) first introduced in \cite{CDP} by framing the loss function as a subgaussian random variable after subtracting its mean. 
This was re-interpreted and relaxed by using Renyi entropy in \cite{BS15} and its authors followed up with the notion zero-CDP (zCDP) in \cite{zCDP}: A mechanism ${\cal M}$ is $\rho$-zCDP if, for all $\alpha> 1$, the Renyi divergence 
$$ {\tt D}_\alpha({\cal M}(d) \| {\cal M}(d')) = \frac{\ln( \mathbb{E}_{o\sim {\cal M}(d)}[e^{(1-\alpha)Z}])}{1-\alpha} \mbox{ with } Z = L^o_{{\cal M}(d)\|{\cal M}(d')}
$$
satisfies
\begin{equation} {\tt D}_\alpha({\cal M}(d) \| {\cal M}(d')) \leq \rho\alpha.
\label{zCDP}
\end{equation}
This DP guarantee requires the tail of $Z$ to be subgaussian, i.e., ${\tt Pr}[Z>t+\rho]<e^{-t^2/(4\rho)}$ for all $t\geq 0$ (the tail behaves like $Z\sim {\cal N}(\rho,2\rho)$). If the loss function satisfies this property for a collection of $k$ mechanisms (each of the mechanisms is $\rho$-zCDP), then their $k$-fold adaptive composition is $k\rho$-zCDP. If a mechanism is $\rho$-zCDP for individual privacy, then it is $g^2\rho$-zCDP for groups of size $g$. This shows that if we can prove that our DP principles lead to a subgaussian tail of the loss function $Z$, then we obtain interpretable DP guarantees even for large $k$ and $g$.

After the introduction of $\rho$-zCDP, Renyi DP (RDP) was introduced by \cite{RDP}; $(\omega,\tau)$-RDP  requires
$$ {\tt D}_\alpha({\cal M}(d) \| {\cal M}(d')) \leq \tau \mbox{ for all } \alpha \in (1,\omega).
$$
Here, $\alpha=1$ bounds the geometric mean of $e^Z$, $\alpha=2$ bounds the arithmetic mean of $e^Z$, $\alpha=3$ bounds the quadratic mean of $e^Z$, etc., and $\alpha=\infty$ bounds the maximum value of $e^Z$ which is equivalent to $(\tau,0)$-DP. RDP also leads to simple computable composition and group privacy. The advantage of zCDP over RDP is that it covers all $\alpha$ at once: Larger $\alpha$ put more weight on the tail of $Z$, also the mean gets larger. This means that $\tau$ in the RDP definition should increase with $\alpha$ and this is realized by zCDP by setting $\tau=\rho \alpha$ for all $\alpha\in (1,\infty)$.

The above discussion leads naturally to the definition of $(\rho,\omega)$-tCDP \cite{tCDP}: A mechanism is $\omega$-truncated $\rho$-CDP if it satisfies (\ref{zCDP}) only for $\alpha\in (1,\omega)$. tCDP requires $Z$ to be subgaussian near the origin (like zCDP), i.e., ${\tt Pr}[Z>t+\rho]<e^{-t^2/(4\rho)}$ for all $0\leq t \leq 2\rho(\omega-1)$, but only subexponential in $Z$'s tail, i.e., we get the weaker subexponential tail bound ${\tt Pr}[Z>t+\rho]\leq e^{(\omega-1)^2\rho}e^{-(\omega-1)t}$.
This relaxes zCDP while still obtaining interpretable DP guarantees for composition and group privacy, and also subsampling.

The main concern with each of the divergence based DP measures is a lack of transparency of how the attacker can best distinguish the hypotheses $o\sim {\cal M}(d)$ and $o\sim {\cal M}(d')$.
The next section introduces the $f$-DP framework which provides a hypothesis testing based approach. It introduces trade-off functions that capture all the information needed for fully characterizing privacy leakage; a trade-off function can be used to derive any divergence based DP guarantee like the ones discussed above (but not the other way around), see Appendix B in \cite{dong2021gaussian}. Rather than extracting a divergence based DP guarantee from a trade-off function for DP-SGD, we will keep the trade-off function itself as it 
%turns out to have a simple form with 
has an easy transparent interpretation.

\section{Gaussian Differential Privacy}
\label{sec:GDP}

Dong et al. \citep{dong2021gaussian} introduced the state-of-the-art DP formulation based on hypothesis testing. From the attacker's perspective, it is natural to formulate the  problem of distinguishing two 
neighboring 
data sets $d$ and $d'$ based on the output of a DP mechanism ${\cal M}$ as a hypothesis testing problem:
$$H_0: \mbox{ the underlying data set is }d \ \ \ \ \mbox{ versus } \ \ \ \  H_1: \mbox{ the underlying data set is }d' .$$
Here, neighboring means that either $|d\setminus d'|=1$ or $|d'\setminus d|=1$. 
More precisely, in the context of mechanism ${\cal M}$, ${\cal M}(d)$ and ${\cal M}(d')$ take as input representations $r$ and $r'$ of data sets $d$ and $d'$ which are `neighbors.' The representations are mappings from a set of indices to data samples with the property that if $r(i)\in d\cap d'$ or $r'(i)\in d\cap d'$, then $r(i)=r'(i)$. This means that the mapping from indices to data samples in $d\cap d'$ is the same for the representation of $d$ and the representation of $d'$. In other words the mapping from indices to data samples for $d$ and $d'$ only differ for indices corresponding to the differentiating data samples in $(d\setminus d')\cup (d'\setminus d)$. In this sense the two mappings (data set representations) are neighbors.

We define the Type I and Type II errors by
$$\alpha_\phi = \mathbb{E}_{o\sim {\cal M}(d)}[\phi(o)]  \mbox{ and } \beta_\phi = 1- \mathbb{E}_{o\sim {\cal M}(d')}[\phi(o)],
$$
where $\phi$ in $[0,1]$ denotes the rejection rule which takes the output of the DP mechanism as input. We flip a coin and reject the null hypothesis with probability $\phi$. The optimal trade-off between Type I and Type II errors is given by the trade-off function
$$ T({\cal M}(d),{\cal M}(d'))(\alpha) = \inf_\phi \{ \beta_\phi \ : \ \alpha_\phi \leq \alpha \},$$ 
for $\alpha \in [0,1]$, where the infimum is taken over all measurable rejection rules $\phi$. If the two hypotheses are fully indistinguishable, then this leads to the trade-off function $1-\alpha$. We say a function $f\in [0,1]\rightarrow [0,1]$ is a trade-off function if and only if it is convex, continuous, non-increasing, 
%at least $0$, 
and $0\leq f(x)\leq 1-x$ for $x\in [0,1]$. 

We define a mechanism ${\cal M}$ to be $f$-DP if $f$ is a trade-off function and
$$
 T({\cal M}(d),{\cal M}(d')) \geq f
$$
for all neighboring $d$ and $d'$.
Proposition 2.5 in \citep{dong2021gaussian} is an adaptation of a result in \citep{wasserman2010statistical} and states that a mechanism is $(\epsilon,\delta)$-DP if and only if the mechanism is $f_{\epsilon,\delta}$-DP, where
$$f_{\epsilon,\delta}(\alpha) =
\min \{ 0, 1-\delta - e^{\epsilon}\alpha, (1-\delta-\alpha)e^{-\epsilon}\}.
$$
We see that $f$-DP has the $(\epsilon,\delta)$-DP  formulation as a special case. It turns out that the original DP-SGD algorithm can be tightly analysed by using $f$-DP.

\subsection{Gaussian DP}
\label{sec:sens}

%\vspace{2mm}
%
%\noindent
%{\bf Gaussian DP:}
In order to proceed, \citep{dong2021gaussian} first defines Gaussian DP as another special case of $f$-DP as follows: We define the trade-off function
$$G_\mu(\alpha) = T({\cal N}(0,1),{\cal N}(\mu,1))(\alpha) = \Phi( \Phi^{-1}(1-\alpha) - \mu ),$$
where $\Phi$ is the standard normal cumulative distribution of ${\cal N}(0,1)$. We define a mechanism to be $\mu$-Gaussian DP if it is $G_\mu$-DP. Corollary 2.13 in \citep{dong2021gaussian} shows  that a mechanism is $\mu$-Gaussian DP if and only if it is $(\epsilon, \delta(\epsilon))$-DP for all $\epsilon\geq 0$, where
\begin{equation} \delta(\epsilon) = \Phi(-\frac{\epsilon}{\mu}+\frac{\mu}{2}) - e^{\epsilon} \Phi(-\frac{\epsilon}{\mu}-\frac{\mu}{2}).
\label{eq:gdp}
\end{equation}

Suppose that a mechanism ${\cal M}(d)$ computes some function $u(d)\in \mathbb{R}^n$ and adds Gaussian noise ${\cal N}(0,(c\sigma)^2{\bf I})$, that is,  the mechanism outputs $o\sim u(d)+{\cal N}(0,(c\sigma)^2{\bf I})$. Suppose that $c$ denotes the sensitivity of function $u(\cdot)$, that is, $$\|u(d)-u(d')\|\leq c$$ 
for neighboring $d$ and $d'$; the mechanism corresponding to one round update in Algorithm \ref{alg:DPs} has {\em sensitivity} $c=2C$. After projecting the observed $o$  onto the line that connects $u(d)$ and $u(d')$ and after normalizing by dividing by $c$, we have that differentiating whether $o$ corresponds to $d$ or $d'$ is in the best case for the adversary (i.e., $\|u(d)-u(d')\|=c$) equivalent to differentiating whether a received output is from ${\cal N}(0,\sigma^2)$ or from  ${\cal N}(1,\sigma^2)$. Or, equivalently, from ${\cal N}(0,1)$ or from  ${\cal N}(1/\sigma,1)$. 
This is how the Gaussian trade-off function $G_{\sigma^{-1}}$ comes into the picture. 

\subsection{Subsampling} \label{sec:sample}

%\vspace{2mm}
%
%\noindent
%{\bf Subsampling:} 
Besides implementing Gaussian noise,   DP-SGD also uses sub-sampling: For a data set $d$ of $N$ samples, ${\tt Sample}_m(d)$   selects a subset of size $m$ from $d$ uniformly at random. We define convex combinations
$$ f_p(\alpha) = p f(\alpha) + (1-p) (1-\alpha)$$
with corresponding $p$-sampling operator 
$$ C_p(f) = \min \{ f_p, f_p^{-1} \}^{**},
$$
where the conjugate $h^*$ of a function $h$ is defined as
$$ h^*(y) = \sup_x \{ yx -h(x) \}$$ 
and the inverse $h^{-1}$ of a trade-off function $h$ is defined as
\begin{equation}
h^{-1}(\alpha) = \inf\{t\in[0,1] \ | \ h(t)\leq \alpha \} \label{inverse}
\end{equation}
and is itself a trade-off function (as an example, we notice that $G_\mu=G_\mu^{-1}$ and we say $G_\mu$ is symmetric).
Theorem 4.2 in \citep{dong2021gaussian} shows that if a mechanism ${\cal M}$ on data sets of size $N$ is $f$-DP, then the subsampled mechanism ${\cal M}\circ {\tt Sample}_{m}$ is $C_{m/N}(f)$-DP.

%Given f-DP and sampling ratio p, how do we come up with fp? It is very
%simple. f is error curve and it is defined as the expectation. Therefore,
%we have the concept
% fp = pf + (1-p)*(1-x) because f is the error curve for cases where we have
%differetenting points and (1-x) is the case we do not have.

The intuition behind operator $C_p$ is as follows. 
First, ${\tt Sample}_m(d)$ samples the differentiating element between $d$ and $d'$ with probability $p$. In this case the computations ${\cal M}\circ {\tt Sample}_{m}(d)$ and  ${\cal M}\circ {\tt Sample}_{m}(d')$ are different and hypothesis testing is possible with trade-off function $f(\alpha)$. With probability $1-p$ no hypothesis testing is possible and we have trade-off function $1-\alpha$. This leads to the convex combination $f_p$. 

Second, we notice if $h=T({\cal M}(d),{\cal M}(d'))$, then $h^{-1}=T({\cal M}(d'),{\cal M}(d))$. Therefore, if ${\cal M}$ is $f$-DP (which holds for all pairs of neighboring data sets, in particular, for the pairs $(d,d')$ and $(d',d)$), then both $h\geq f$ and $h^{-1}\geq f$ and we have a symmetric upper bound $\min\{h,h^{-1}\}\geq f$. Since $f$ is a trade-off function, $f$ is convex and we can compute a tighter upper bound: $f$ is at most the largest convex function $\leq \min\{h,h^{-1}\}$, which is equal to the double conjugate $\min\{h,h^{-1}\}^{**}$. From this we obtain the definition of operator $C_p$.

%Why do we have fp^-1? It is because T(P,Q) and T(Q,P) are the same and
%thus, the f-DP should be symmetric. By definition of f-DP, f does not need
%to be symmetric. But we are interested in the tight bound, then thus, we
%want to have symmetric property and thus fp^-1 comes into the picture. And
%again because we want to have the tightbound, we need min(fp,fp^-1). But
%min(fp,fp^-1) is not convex we need to derive a tight and convex function
%from min(fp,fp^-1). How to do it? Since min(fp,fp^-1) is not convex, we
%have to apply the conjugate operation twice to get it, i.e., this is why we
%need min(fp,fp^-1)**.

%= Note that we do not apply max(fp,fp^-1) because in case of subsampling,
%min(fp,fp^-1) gives us a better bound.

\subsection{Composition}

%\vspace{2mm}
%
%\noindent
%{\bf Composition:}
The tensor product $f\otimes h$ for trade-off functions $f=T(P,Q)$ and $h=T(P',Q')$ is well-defined by 
$$f\otimes h = T(P\times P',Q\times Q').$$
Let $y_i \leftarrow {\cal M}_i(\texttt{aux},d)$ with $\texttt{aux}=(y_1,\ldots, y_{i-1})$. Theorem 3.2 in \citep{dong2021gaussian} shows that if ${\cal M}_i(\texttt{aux},.)$ is $f_i$-DP for all $\texttt{aux}$, then the composed mechanism ${\cal M}$, which applies ${\cal M}_i$ in sequential order from $i=1$ to $i=T$, is 
$(f_1\otimes \ldots \otimes f_T)$-DP.
%$f^{\otimes T}$-DP.
The tensor product is commutative.

As a special case Corollary 3.3 in \citep{dong2021gaussian} states that composition of multiple Gaussian operators $G_{\mu_i}$ results in $G_{\mu}$ where
$$
\mu=\sqrt{\sum_i \mu_i^2}.
$$
%$\mu=(\sum_i \mu_i^2)^{1/2}$.

\subsection{Tight Analysis of DP-SGD}

%\vspace{2mm}
%
%\noindent
%{\bf Tight Analysis DP-SGD:}
We are now able to formulate the differential privacy guarantee of original DP-SGD 
%(which implements subsampling, individual clipping, and mini-batch SGD) 
since it is a composition of  subsampled Gaussian DP mechanisms. Theorem 5.1 in \citep{dong2021gaussian} states that DP-SGD 
%(with subsampling, individual clipping and mini-batch SGD) 
as introduced in \cite{abadi2016deep} 
%with (the slightly more general) update $U$ defined in (\ref{eq:ind}) before adding Gaussian noise
%in Algorithm \ref{alg:??} [TO DO: work with $2\sigma$ in pseudo codes as in $f$-DP paper] 
is
%[TO DO -- we now use $|S_b|=ms$ instead of $s$ here]
%\footnote{TO DO: In appendix explain where DP-SGD paper goes wrong. Their DP-SGD algorithm uses noise ${\cal N}(0,C^2(2\sigma)^2 {\bf I})$ compared to ${\cal N}(0,C^2\sigma^2 {\bf I})$ in our version of the DP-SGD algorithm.}
$$ C_{m/N}(G_{\sigma^{-1}})^{\otimes T}\mbox{-DP},$$
where $T=(N/m)\cdot E$ is the total number of local rounds. 
%[TO DO: Use another symbol. Maybe $R$?]
Since each of the theorems and results from \citep{dong2021gaussian} enumerated above are exact, we have a tight analysis.
%(in that there does not exist a trade-off function $f(\alpha)$ closer to $1-\alpha$ such that DP-SGD is $f$-DP).
%\textbf{Ha's note: what does it mean in that there does not exist a trade-off function $f(\alpha)$ closer to $1-\alpha$ such that DP-SGD is $f$-DP?}
This leads in \citep{zhu2021optimal} to a (tight) differential privacy accountant\footnote{The tight analysis has,  cited from \citep{dong2021gaussian}, ``the disadvantage
is that the expressions it yields are more unwieldy: they are computer evaluable, so usable in
implementations, but do not admit simple closed form." For this reason we need an accountant method.} (using complex characteristic functions for each of the two hypotheses based on taking  Fourier transforms), which can be used by a client to keep track of its current DP guarantee and to understand when to stop  helping the server to learn a global model. Because the accountant is tight, it improves over the momentum accountant method of \cite{abadi2016deep}.

\subsection{Strong Adversarial Model}
\label{sec:strong}

%data set size assumption, remove factor 2

%asymptotics

%\vspace{2mm}
%
%\noindent
%{\bf Strong Adversarial Model:}
%In the proof of our main theorem in Appendix \ref{sec:DPproof} we define the adversary under which tightness holds. 
We assume an adversary who knows the differentiating samples in $d\setminus d'$ and $d'\setminus d$,  but who a-priori (before mechanism ${\cal M}$ is executed) may only know (besides say a 99\% characterization of $d
\cap d'$) an estimate of the number of samples in the intersection of $d$ and $d'$, i.e., the adversary knows $|d\cap d'|+noise$  where the noise is large enough to yield a `sufficiently strong' DP guarantee with respect to the size of the used data set ($d$ or $d'$). Since ${\cal M}$ does not directly reveal the size of the used data set, we assume (as in prior literature) that the effect of $N=|d|\neq N'=|d'|$ contributes at most a very small amount of privacy leakage, sufficiently small to be discarded in our DP analysis: That is, we may as well assume $N=N'$ in our DP analysis. 

This means that the tight $f$-DP analysis of DP-SGD holds, even if we use  the definition of neighboring data sets stating that either $|d\setminus d'|=1$ or $|d'\setminus d|=1$; the original $f$-DP analysis considers the case $|d\setminus d'|=|d'\setminus d|=1$ and this requires the factor $2C\sigma$ in Algorithm \ref{alg:DPs}. If we assume no knowledge about the exact data set sizes (as discussed above) and if we assume probabilistic sampling (see Section \ref{sec:strong}), then we may only use $C\sigma$ saving a factor 2 (which helps convergence to a higher accuracy). 

%of \cite{abadi2016deep} where either $|d\d'$

In the setting of $N=N'$ the  DP analysis in prior work  considers an adversary ${\cal A}dv$ who can mimic mechanism ${\cal M}\circ {\sf Sample}_m$ in that it can replay into large extent how ${\sf Sample}_m$ samples the used data set ($d$ or $d'$): 
%For each round, the DP analysis considers two cases, either (a)
We say a round has $k$ differentiating data samples if
${\sf Sample}_m$ sampled a subset of indices which contains exactly $k$ indices of  differentiating data samples from $(d\setminus d')\cup (d'\setminus d)$. 
The adversary knows how ${\sf Sample}_m$ operates and can derive a joint probability distribution $\mathbb{P}$ of the number of differentiating data samples 
for each round within the sequence of rounds that define the series of epochs during which updates are computed.
%for the sequence of rounds within the epochs during which updates are computed. 
%We consider two types of strong adversaries in our proofs when bounding trade-off functions:
%in our DP analysis: 

Adversary ${\cal A}dv$ does not know the exact instance drawn from $\mathbb{P}$ but  is, in the DP proof, given the ability to realize for each round the trade-off function $f_k(\alpha)$ that corresponds to hypothesis testing between  ${\cal M}\circ {\tt Sample}_{m}(d)$ and  ${\cal M}\circ {\tt Sample}_{m}(d')$ if ${\tt Sample}_{m}$ has selected $k$ differentiating samples in that round. 
In the DP analysis that characterizes $f_k(\alpha)$, adversary ${\cal A}dv$ is given  knowledge about the mapping from indices to values in $d$ or $d'$. Here (as discussed before), the mapping from indices to values in $d\cap d'$ is the same for the mapping from indices to values in $d$ and the mapping from indices to values in $d'$. 
Furthermore, the adversary  
can replay how ${\sf Sample}_m$  samples a subset of $m$ indices from\footnote{By assuming $N=N'$ in the DP analysis, knowledge of how ${\sf Sample}_m$ samples a subset of indices  cannot be used to differentiate the hypotheses of $d$ versus $d'$ based on their sizes (since the index set corresponding to $d$ is exactly the same as the index set corresponding to $d'$).} $\{1,\ldots, N=N'\}$, and it knows all the randomness used by ${\cal M}$ before ${\cal M}$ adds Gaussian noise for differential privacy (this includes when and how the interrupt service routine overwrites the local model). This strong adversary represents a worst-case scenario for the `defender' when analyzing the differential privacy of a single round. 
For DP-SGD this analysis for neighboring data sets 
leads to the argument of Section \ref{sec:sample} where with probability $p$ (i.e., $k=1$) the adversary can achieve trade-off function $f(\alpha)$ and with probability $1-p$ (i.e., $k=0$) can achieve trade-off function $1-\alpha$ leading ultimately to operator $C_p$. This in turn leads to the trade-off function  $C_{m/N}(G_{\sigma^{-1}})^{\otimes T}$ with $p=m/N$,
%as discussed above, 
which is {\em tight for adversary ${\cal A}dv$}. 
Although usually not explicitly stated, we notice that adversary ${\cal A}dv$ is used in DP analysis of current literature including the moment accountant method of \cite{abadi2016deep} for analysing $(\epsilon,\delta)$-DP and analysis of divergence based DP measures.

In the DP analysis  adversary ${\cal A}dv$ is given knowledge about the number $k$ of differentiating samples when analysing a single round. That is, it is given an instance of $\mathbb{P}$ projected on a single round. We notice that in expectation the sensitivity (see Section \ref{sec:sens}) of a single round as observed by adversary ${\cal A}dv$ for neighboring data sets is equal to $(1-p)\cdot 0 + p\cdot 2C=(m/N)\cdot 2C$ and this gives rise to an `expected' trade-off function $G_{1/(\sigma N/m)}$. Composition over $c^2 (N/m)^2$ rounds gives $G_{c/\sigma}$. This leads us to believe that $C_{m/N}(G_{\sigma^{-1}})^{\otimes T}$ converges to $G_{c\cdot h(\sigma)}$ for $T=c^2 (N/m)^2 \rightarrow \infty$ 
%\textcolor{red}{use f-DP eprint}
(or, equivalently, $\sqrt{T}\cdot m/N =c$ with $T\rightarrow \infty$ and $N\rightarrow \infty$) where $h(\sigma)$ is some function that only depends on $\sigma$ (see below). This intuition is confirmed by Corollary 5.4 in \cite{dong2021gaussian}, and is also indirectly confirmed by \cite{Dijk2021ProactiveDP} which shows that DP-SGD is $(\epsilon,\delta)$-DP for $\sigma = \sqrt{2(\epsilon +\ln(1/\delta))/\epsilon}$  for a wide range of parameter settings $N,m,T$ with $T$ at most $\approx \epsilon (N/m)^2/2$, and which matches Corollary 5.4 in \cite{dong2021gaussian} in that
the upper bound on $T$ can at most be a constant factor $\approx 8$ larger (without violating the corollary).

By using $T=(N/m)\cdot E$, we have convergence to $G_{c\cdot h(\sigma)}$ for $c=\sqrt{mE/N}$ with $E\rightarrow \infty$ and $N\rightarrow \infty$. By using a Taylor series expansion we have  %\cite{dong2021gaussian}
$$h(\sigma) = \sqrt{2(e^{\sigma^{-2}}\Phi(\frac{3\sigma^{-1}}{2})+3\Phi(-\frac{\sigma^{-1}}{2})-2)}
= \sigma^{-1} \cdot \sqrt{1-\frac{5}{8} \frac{\sigma^{-1}}{\sqrt{2\pi}}+O(\sigma^{-2})}.
$$
This shows that for `large $N$ and $E$'  DP-SGD is approximately $G_\mu$-DP with $\mu=\sqrt{mE/(N\sigma^2)}$. Notice that $\sigma$ cannot be too small otherwise $h(\sigma)$ behaves like $e^{1/(2\sigma^2)}$ which yields a very weak DP guarantee.

Clearly, a weaker (than ${\cal A}dv$) adversary with less capability (less knowledge of the used randomness by ${\sf Sample}_m$ and ${\cal M}$) achieves a trade-off function $\geq C_{m/N}(G_{\sigma^{-1}})^{\otimes T}$ closer to the ideal $1-\alpha$. It remains an open problem to characterize realistic weaker adversaries that lead to larger (lower bounds of) trade-off functions. %The results in this chapter, like previous work, assumes the strong adversary in DP analysis.

%%TO DO: knows ISR contribution ..

%%trade-off function  $C_{m/N}(G_{\sigma^{-1}})^{\otimes T}$

%%[TO DO: explain DP accountant, cite their papers. Compare to momentum accountant.]

\subsection{Group Privacy}

%\vspace{2mm}
%
%\noindent
%{\bf Group Privacy:} 
Theorem 2.14 in \citep{dong2021gaussian} analyzes how privacy degrades if $d$ and $d'$ do not differ in just one sample, but differ in $g$ samples. If a mechanism is $f$-DP, then it is $$[1-(1-f)^{\circ g}]\mbox{-DP}$$ for groups of size $g$ (where $\circ g$ denotes the $g$-fold iterative composition of function $1-f$, where $1$ denotes the constant integer value $1$ and not the identity function, i.e., $(1-f)(\alpha)=1-f(\alpha)$). This is a tight statement in that {\em there exist} $f$ such that the trade-off function for groups of size $g$ cannot be bounded better. In particular, for $f=G_\mu$ we have $G_{g\mu}$-DP for groups of size $g$. 

The intuition behind the $[1-(1-f)^{\circ g}]$-DP result is that the adversary can create a sequence of data sets $d_0=d$, $d_1$, \ldots, $d_{g-1}$, $d_{g}=d'$ such that each two consecutive data sets $d_i$ and $d_{i+1}$ are neighboring. We know that $T({\cal M}(d_i),{\cal M}(d_{i+1}))\geq f$. 
%By definition, $f(\alpha)$ is a lower bound on the Type I vs Type II error curve. Hence, $1-f(\alpha)$  upper bounds the cloud of 
For each rejection rule we may plot a point (in x and y coordinates) $$(\mathbb{E}_{o\sim {\cal M}(d_i)}[\phi(o)], \ \mathbb{E}_{o\sim {\cal M}(d_{i+1})}[\phi(o)]).$$ 
Since $f(\alpha)$ is a lower bound on the Type I vs Type II error curve, the resulting collection of points is upper bounded by the curve $1-f(\alpha)$.
%(y-axis) versus $\mathbb{E}_{o\sim {\cal M}_i(d)}[\phi(o)]$ (x-axis). That is,
We have that $\alpha=\mathbb{E}_{o\sim {\cal M}(d_i)}[\phi(o)]$ is mapped to $$\mathbb{E}_{o\sim {\cal M}(d_{i+1})}[\phi(o)]\leq 1-f(\alpha)=(1-f)(\alpha).$$ By transitivity, we have that $\alpha=\mathbb{E}_{o\sim {\cal M}(d=d_0)}[\phi(o)]$ is mapped to $$\mathbb{E}_{o\sim {\cal M}(d'=d_{g})}[\phi(o)]\leq (1-f)^{\circ g}(\alpha).$$ This yields the lower bound $$T({\cal M}(d),{\cal M}(d'))\geq 1-(1-f)^{\circ g}$$ on the Type I vs Type II error curve.

%Talk about strong adversary and tightness -- 
Let $\phi[\alpha]$ denote a rejection rule that realizes the mapping from $$\alpha=\mathbb{E}_{o\sim {\cal M}(d_i)}[\phi[\alpha](o)] \ \ \mbox{ to } \ \ (1-f)(\alpha)=\mathbb{E}_{o\sim {\cal M}(d_{i+1})}[\phi[\alpha](o)].$$ Then  
the mapping from $(1-f)^{\circ i}(\alpha)=\mathbb{E}_{o\sim {\cal M}(d_i)}[\phi(o)]$ to $(1-f)^{\circ (i+1)}(\alpha)=\mathbb{E}_{o\sim {\cal M}(d_{i+1})}[\phi(o)]$ is realized by $\phi=\phi[(1-f)^{\circ i}(\alpha)]$.
This shows that the lower bound $1-(1-f)^{\circ g}$ is tight only if we can choose all $\phi[(1-f)^{\circ i}(\alpha)]$ equal to one another.
In general, this may not be the case for DP-SGD and future work may be able to produce an improved analysis for DP-SGD.
%\textcolor{red}{to improve ..}  
%the class of mechanisms defined by our algorithmic framework 
%%% REMOVE %%%%
%for which it turns out that this lower bound is not tight at all; rather than a multiplicative factor $g$ as in the mentioned $G_{g\mu}$-DP guarantee we have a $\sqrt{g}$ dependency for  adversary ${\cal A}_1$ \cite{vanDijk2022dp} (and this should also hold for the seemingly weaker adversary ${\cal A}_0$). This is done by considering how, due to sub-sampling, the $g$ differentiating samples are distributed across all the rounds within an epoch and how composition of trade-off functions across rounds yields the $\sqrt{g}$ dependency.
%We accomplish this by analyzing $T({\cal M}(d),{\cal M}(d'))$ directly and consider how a group of $g$ differentiating samples affect sensitivity
%This is because 

%Talk about that we will consider: distributions ..

\subsection{DP-SGD's Trade-Off Function}
\label{sec:trade}

%\textcolor{red}{Rewrite: use the $f$-DP approximation, include composition over groups ($p\approx gm/N$, sensitivity is $g$ larger hence $\sigma$ is $g$ smaller, $mE/N = (gm/N)E/g$, so we have $\sqrt{g}$ contribution}
%Assuming adversary ${\cal A}_1$, recent work \cite{vanDijk2022dp} shows that 
%for $g=1$, DP-SGD is $h$-DP for
%$$ h \approx G_{\sqrt{(1+1/\sqrt{2E})E}/\sigma}
%$$
%and if DP-SGD is $h$-DP, then it is upper bounded by $h\leq \bar{h}$ with
%$$
%\bar{h} \approx G_{\sqrt{(1-1/\sqrt{2E})E}/\sigma}.
%$$
%The approximations become tight if $e^{-E}$ tends to zero. 
%%%%%%%%%%%%%%%%
%%%%%%%%%%%%%%%%
%This result shows that the $C_{m/N}(G_{\sigma^{-1}})^{\otimes T}$-DP guarantee with $T=(N/m)\cdot E$ discussed above has the property
%$$ h\leq C_{m/N}(G_{\sigma^{-1}})^{\otimes (N/m)\cdot E} \leq \bar{h},
%$$
%which shows 
%Notice that we do not need to compute the subsampling operator (which is only specific to adversary ${\cal A}_0$) and composition tensor in order to get a good approximation. The approximation is easy to interpret as it behaves like $G_{\sqrt{E}/\sigma}$ as opposed to 
%general $f$-DP theory which has,  cited from \citep{dong2021gaussian}, ``the disadvantage
%is that the expressions it yields are more unwieldy: they are computer evaluable, so usable in
%implementations, but do not admit simple closed form." 

We remind the reader that we can  directly infer $(\epsilon,\delta)$-DP guarantees from (\ref{eq:gdp}); function $\delta(\epsilon)$ turns out to be completely independent from the data set size $N$, hence, see Section \ref{section:epsdelta}, setting $\delta(\epsilon)=1/N$ favorably biases smaller data sets. Appendix B in \cite{dong2021gaussian} shows how to infer divergence based DP guarantees. In particular, $G_{\mu}$-DP implies $(\omega,\frac{1}{2\mu^2}\cdot \omega)$-RDP (Renyi differential privacy) for any $\omega> 1$, hence, we have
$$\frac{1}{2\mu^2}\mbox{-zCDP}.$$
For group privacy with $g\geq 1$, we have $G_{\mu\cdot g}$-DP and RDP and zCDP scale with another factor $g^2$.

%As noted in \cite{vanDijk2022dp}  the resulting 
%$G_{\sqrt{E}/\sigma}$-DP guarantee for individual privacy scales with the square root of the total number $E$ of epochs, and does not depend on the explicit number of rounds executed within these epochs. In other words, even though more local updates can be observed by the adversary, this turns out not to lead to more privacy leakage. This is because each local update is based on a smaller batch of training data samples which in itself leads to less privacy leakage per local update. 
%We remind the reader about our discussion in Section \ref{sec:princ} where we show that the role of  mini-batch size $m$ (and, hence, the total number of rounds $N/m$ per epoch) is implicitly captured in $\sigma$.

%For group privacy with $g\geq 1$,  \cite{vanDijk2022dp} shows a similar approximate DP guarantee as for $g=1$ where everywhere $E$ is substituted by $gE$; we have that DP-SGD with sampling based on `shuffling' is $G_{\sqrt{gE}/\sigma}$-DP for groups of size $g$.
%%differentiating samples. 
%This leads to a $\sqrt{gE}$ dependency and not a linear dependency in $g$ (and notice that we obtain $gE/(2\sigma^2)$-zCDP with a linear dependency on $g$ rather than $g^2$). 
%%%%%%%%%%%%

%our result

%(In our main theorem we will consider the general case $g=\max\{ |d\setminus d'|, |d'\setminus d|\}$ in order to analyse `group privacy.') 

%\subsection{DP Accountants}

\section{Future Work}\label{sec:future}

We are still in the midst of bringing DP-SGD to practice where we want to achieve good convergence to and accuracy of the final global model and where we have a strong DP guarantee (the trade-off function should be close to $1-\alpha$ which represents random guessing between the two hypotheses). Towards finding a good balance between utility and privacy, we discuss a couple future directions in next subsections.

\subsection{Using Synthetic Data}

One main problem is that local data is used for training models for various learning tasks. Each application of DP-SGD will leak privacy since the local data set is being re-used. 
%Without any pre-procesing, if we want 
One way to control and be in charge of the amount of privacy leakage is to have data samples in local client data expire according to some expiration date (per sample). This is problematic because in our current data economy, data is a valuable asset which we do not want to give a limited lifetime.

In order to cope with this problem, a client may decide to not use its own local data set in each of these DP-SGD 
instantiations. Instead, differential private GAN~\citep{goodfellow2020generative} modeling can be used to learn a distribution model based on a local data set that generates synthetic data with a similar distribution. Due to the post-processing lemma, we can freely use the synthetic data in any optimization algorithm and FL approach. This circumvents multiple use of DP-SGD, but requires the design of differential private GAN which produces `high' quality synthetic data. This is an open problem: GAN modeling is itself a learning task which can use the DP-SGD approach for the discriminator (which is very noise sensitive). Here, we use DP-SGD only once and as soon as a GAN model is learned, it can
%The GAN model can even 
be published and transmitted to the central server who uses the GAN models from all clients to generate synthetic samples on which it trains a global model for a learning task of its choice. Of course, as a caveat, working with synthetic data may not lead to a global model with good test accuracy on real data. Notice that by using synthetic data we avoid the FL paradigm altogether since the large amounts of data distributed over clients is now compressed into (relatively short transmittable) representations that code GAN models. 

In the same line of thinking, if differentially private GAN models do not lead to high quality synthetic data, then we will want to research other general methods for pre-processing local data that filter or hide features that are considered privacy sensitive. This brings us back to the basics of how a membership or inference attack is actually implemented in order to understand what type of information should be filtered out for making reconstruction of certain types of private data hard or unreliable.

%\cite{some work by alex pentland?}

%re-using a data set ... expiry date on data samples (but data is money)

\subsection{Adaptive Strategies}

We need to fine tune parameters and this can be done during DP-SGD's execution: Consecutive segments of multiple rounds may work with their own $m$, $\sigma$, and $C$. Into what extent does an adaptive approach work, where the current convergence rate and test accuracy (preferably based on public data at the server so as not to leak additional privacy) of the current global model is used to determine $(m,\sigma,C)$ for the next segment?  

In Section \ref{sec:princ} we discussed the benefit of adaptive reducing the clipping constant $C$ (based on prior rounds or based on using a DP approach within a round to collect information that influences the choice of the used $C$ in that round). Similarly, since a smaller $\sigma$ directly reduces the amount of noise added to the global model and therefore increase the final accuracy, it makes sense to reduce $\sigma$ once convergence has been achieved. After reducing $\sigma$, new convergence  to an improved global model may start. The problem is that a lower $\sigma$ leads to more privacy leakage. For this reason we want to lower $\sigma$ to a smaller $\hat{\sigma}$ only for e.g. the final epoch. 

%; \textcolor{red}{remove next part} this yields for adversary ${\cal A}_1$ a trade-off function approximately equal to
%$$G_{\sqrt{g(E-1)}/\sigma}\otimes G_{\sqrt{g}/\hat{\sigma}}=
%G_{\sqrt{g}\cdot \sqrt{(E-1)/\sigma^2 + 1/\hat{\sigma}^2}}.
%$$
%We may decide to  choose a significantly smaller $\hat{\sigma}=\sigma/\sqrt{E+1}$ which yields $G_{\sqrt{2gE}/\sigma}$-DP, sacrificing a factor $\sqrt{2}$ in the differential privacy guarantee. The significantly smaller $\hat{\sigma}$ will likely improve the accuracy of the final global model during the last epoch. Notice that, when adapting $C$ and $\sigma$, it also makes sense to fine tune $m$ accordingly as the sensitivity to a lack of information dispersal may reduce for smaller $C$ and $\sigma$. 

We may also modify the noise distribution: DP-SGD selects noise $N$ from a Gaussian distribution. before adding $N$ to the round update, we may replace $N$ by $a\cdot {\tt arsinh}(a^{-1}\cdot N)$ where ${\tt arsinh}(x)=\ln (x+\sqrt{x^2+1})$ as suggested for tCDP \cite{tCDP}. The result resembles the same Gaussian but with exponentially faster tail decay and this may help improving the convergence to and accuracy of the final global model. Here, we notice that for the same reason of faster tail decay, DP-SGD chooses to use Gaussian noise over Laplace noise.

%\textcolor{red}{study adaptive clipping with local SGD in batch clipping and batch normalization etc.} 
Finally, DP-SGD can be placed in a larger algorithmic framework with  DP guarantees for a more general clipping strategy (including clipping a batch of gradients) which allows more general optimization algorithms (beyond mini-batch SGD), and more general sampling strategies (in particular a sampling strategy based on `shuffling'). 
%\cite{vanDijk2022dp}.

It remains an open problem to unveil adaptive strategies possibly in a more general algorithmic framework that optimally  balance utility and differential privacy. Here we prefer to discover adaptive strategies that proactively provide the DP guarantee based on changed parameter settings, i.e., we do not want to change parameters based solely on utility and discover later (by using a differential privacy accountant) that this has violated or is about to  violate our privacy budget.

%\subsection{A More General Algorithmic Framework}

%by
%still dominated by $\sqrt{E}/\sigma$ 

%fine tuning parameters: sequence with $m,s,\sigma,C$, all adaptive, different segments can be analyzed as above, and combined using accountant, client in control whether it wants to participate if selected bu server; proactive SGD

%other noise distribution, not Gaussian but e.g. Gaussian noise $N$ replaced by $a\cdot {\tt arsinh}(a^{-1}\cdot N)$ where ${\tt arsinh}(x)=\ln (x+\sqrt{x^2+1})$ as suggested for tCDP \cite{tCDP}. This is the same Gaussian but with exponentially faster tail decay and this may help improving the convergence to and accuracy of the final global model. For the same reason of faster tail decay, we also prefer to use Gaussian noise over Laplace noise.

%better trade-off privacy leakage versus utility/accuracy

\subsection{DP Proof: A Weaker Adversarial Model}

Section \ref{sec:strong} explains the strong adversarial model used in DP analysis under which the derived DP guarantee is tight. In practice, this is too strong. In general, we may assume a weaker adversary with less capability in terms of knowledge about the used randomness by ${\tt Sample}_m$ and ${\cal M}$. By explicitly stating the knowledge of a weaker adversary in combination with assumptions on the data set itself, we may be able to derive an $f$-DP guarantee with $f(\alpha)$ closer to $1-\alpha$. It remains an open problem to exploit such a line of thinking.

\subsection{Computing Environment with Less Adversarial Capabilities}

%transform training data into a set of DP-protected synthetic data: unlimited use of the synthetic training data set, it can even be published

%weaker adversarial model in proof

In order to impose restrictions on  adversarial capabilities we may be able to use confidential computing techniques such as  secure processor technology \cite{IntelSGX}, homomorphic computing with secret sharing and/or secure Multi-Party Computation (MPC) \cite{crypten}, and possibly even hardware accelerated fully homomorphic encryption \cite{Gentry09,feldmann2021f1}; for a survey see \cite{kairouz2021advances,li2020federated}.
These techniques hide round updates in encrypted form. Hence, only the final global model itself (if it is published) or querying the final global model (if it is kept private) can leak information about how local data sets shaped the final model. This means that CDP, see Section \ref{sec:DP}, is still needed. CDP has a better trade-off between privacy and utility compared to LDP as discussed in this chapter. However, confidential computing does not come for free: Either we need to assume a larger Trusted Computing Base (TCB) in the form of trusted hardware modules or processors at the clients, intermediate aggregators, and server   or we need a Trusted Third Party (TTP). E.g., in the secure MPC solution of \cite{crypten} the generation of Beaver triples is outsourced to a TTP otherwise impractical additional communication among clients and server is needed (for an oblivious transfer phase in MPC). We are still studying  balanced and practical combinations of confidential computing techniques including the use of differential privacy.